\documentclass[10pt,twocolumn,letterpaper]{article}

\usepackage{cvpr}
\usepackage{times}
\usepackage{epsfig}
\usepackage{graphicx}
\usepackage{amsmath}
\usepackage{amssymb}

\usepackage[pagebackref=true,breaklinks=true,letterpaper=true,colorlinks,bookmarks=false]{hyperref}

\cvprfinalcopy 

\def\cvprPaperID{8639} 

\ifcvprfinal\fi
\begin{document}

\title{Hierarchical Attention Networks for Medical Image Segmentation}

\author{Fei Ding$^1$, Gang Yang\thanks{Gang Yang is the corresponding author (yanggang@ruc.edu.cn).}$^{*1}$, Jinlu Liu$^3$, Jun Wu$^4$, Dayong Ding$^2$, Jie Xv$^5$, Gangwei Cheng$^6$, Xirong Li$^1$\\
\\
$^1$AI \& Media Lab, School of Information, Renmin University of China\\
$^2$Vistel AI Lab, Visionary Intelligence Ltd.\\
$^3$AInnovation Technology Co., Ltd.\\
$^4$Northwestern Polytechnical University\\
$^5$Beijing Tongren Hospital\\
$^6$Peking Union Medical College Hospital
}
\maketitle

\begin{abstract}
The medical image is characterized by the inter-class indistinction, high variability, and noise, where the recognition of pixels is challenging. Unlike previous self-attention based methods that capture context information from one level, we reformulate the self-attention mechanism from the view of the high-order graph and propose a novel method, namely Hierarchical Attention Network (HANet), to address the problem of medical image segmentation. Concretely, an HA module embedded in the HANet captures context information from neighbors of multiple levels, where these neighbors are extracted from the high-order graph. In the high-order graph, there will be an edge between two nodes only if the correlation between them is high enough, which naturally reduces the noisy attention information caused by the inter-class indistinction. The proposed HA module is robust to the variance of input and can be flexibly inserted into the existing convolution neural networks. We conduct experiments on three medical image segmentation tasks including optic disc/cup segmentation, blood vessel segmentation, and lung segmentation. Extensive results show our method is more effective and robust than the existing state-of-the-art methods.
\end{abstract}
\section{Introduction}
Semantic segmentation plays a critical role in various computer vision tasks such as image editing, automatic driving, and medical diagnosing. It has been well handled by the powerful methods driven by Convolutional Neural Networks (CNNs). CNNs have achieved promising results in image recognition, where a key operation is the global pooling. The global pooling makes the feature representation of an image contain global context information. However, pixel features only contain local context information, which makes the recognition of pixels challenging. It is more challenging to distinguish the confusing pixels in the medical images than other kinds of images, because of the inter-class indistinction, high variability, and noise.
\begin{figure}[t]
\centering
\includegraphics[width=0.41\textwidth]{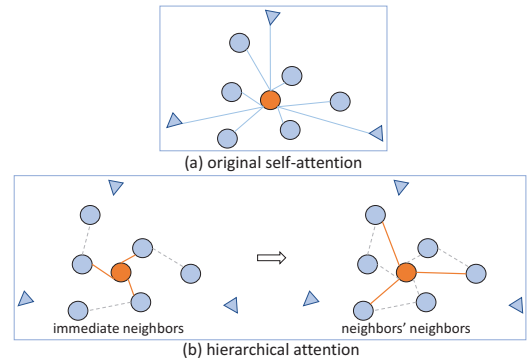}
\caption{Diagrams of original self-attention and our two-levels hierarchical attention. In original self-attention (a), the center node (orange circle) is influenced by its whole neighbors, including the nodes from a different class (triangle). 
In our hierarchical attention (b), we search N-degree neighbors for the center node from the sparse graph (dotted lines are edges) where there will be an edge between two nodes only if the correlation between them is high enough.
In the left part of our hierarchical attention (b), the center node is influenced by its immediate neighbors, and in the right part of (b), the center node is influenced by its neighbors' neighbors, and the information of two levels will be fused to enrich the context information.}
\label{fig:intro}
\end{figure} 
Various pioneering fully convolutional network (FCN) approaches have taken into account the contexts to improve the performance of semantic segmentation. The U-shaped networks~\cite{U-net, r2u-net} achieve promising results on medical image segmentation, which enable the use of rich context information. Other methods exploit context information by the dilated convolutions~\cite{deeplabv3,deeplabv3+,denseaspp,ce-net} or multi-scale pooling \cite{pspnet,poolnet}.
The above methods have revealed that CNNs with large and multi-scale receptive fields are more possible to learn translation-invariant features. However, the fact that utilizing the same weights of feature detectors at all locations does not satisfy the requirement that different pixels need different contextual dependencies. Therefore, many self-attention based approaches~\cite{non-local,transformer} are proposed, which focus on aggregating context information in an adaptive manner. In the self-attention, for the feature at a certain position, it is updated via aggregating features at all other positions in a weighted sum manner, where the weights are decided by the similarity between two corresponding features. However, two issues will reduce the performance when applying the self-attention to realize medical image segmentation. Firstly, the weighted summed features contain information about other categories (see Figure \ref{fig:intro} (a)), which means that attention map contains noise, and this kind of noise will increase dramatically in the medical image with inter-class indistinction.  The above view is proven in our experiments and shown in Figure \ref{fig:map}. Secondly, CNNs are powerful for their ability to generate a hierarchical object representation~\cite{lecun}, which is related to the human visual system. However, the existing self-attention methods only aggregate context information of one level which we term one-level attention, and they can not learn a hierarchical attention representation. Because of the importance of semantic segmentation in medical image analyzing, introducing the hierarchical attention mechanism into the medical image domain to generate more powerful pixel-wise feature representations will yield a general advantage. 

Based on the above observation, we reformulate the original self-attention mechanism from the view of the high-order graph and propose a novel encoder-decoder structure for medical image segmentation, called Hierarchical Attention Network (HA-Net). Instead of one-level information mixing, HANet realizes an effective aggregation of hierarchical context information. Specifically, we first compute the initial attention map that represents the similarities between two corresponding features. Then, the initial attention map is utilized to extract a sparse graph, where there will be an edge between two nodes only if the correlation between them is high enough. Further, we search N-degree neighbors for each node from the sparse graph (see Figure~\ref{fig:intro} $\left(b\right)$). Finally, the nodes are updated via mixing information of neighbors at various distances. Our hierarchical attention mechanism embedded in the HANet focuses on optimizing the neighbor relations only with high confidence and aggregates latent information of hierarchical neighbors. It naturally reduces noisy attention information caused by the inter-class indistinction of medical images and can generate a wide class of feature representations.

The main contributions of this paper are listed as follows:
\begin{itemize}
  \item
   We reformulate the self-attention mechanism in a high-order graph manner, which aggregates hierarchical context information to enhance the discriminative ability of feature representations. To the best of our knowledge, this paper is the first to introduce the high-order graph into the self-attention mechanism. 
  \item
  We build the proposed hierarchical attention mechanism as a flexible module for neural networks, yielding powerful compact attention architectures for medical image segmentation.
  \item
  Extensive experiments on four datasets, including REFUGE dataset\footnote{\url{https://refuge.grand-challenge.org/}\label{REFUGE}}, Drishti-GS1 dataset \cite{DGS}, DRIVE dataset \cite{DRIVE}, and LUNA dataset\footnote{\url{https://www.kaggle.com/kmader/finding-lungs-in-ct-data/data/}\label{LUNA}}, demonstrate the superiority of our approach over existing state-of-the-art methods. The code will be released.
\end{itemize}
\section{Related works}
\textbf{Semantic segmentation.} Fully convolutional network (FCN) \cite{fcn} based methods have made great progress in semantic segmentation, which attempt to predict pixel-level semantic labels of a given image. Recently, several model variants are proposed  to enhance the context information aggregation for semantic segmentation. Atrous spatial pyramid pooling (ASPP) based methods \cite{deeplabv2,deeplabv3,deeplabv3+,denseaspp} aggregate context information via parallel dilated convolutions \cite{overfeat, dilated_conv, local-global}. Other methods \cite{pspnet,poolnet} collect contextual information of different scales via multi-scale pooling. The above methods reveal the advantage of large and multi-scale receptive fields in semantic segmentation. Moreover, the U-shaped networks \cite{U-net, r2u-net, ET-Net, dunet, deu-net} enable multi-level feature extraction and successive feature aggregation, which have achieved promising results on medical image segmentation. The U-shaped networks work with a few training samples and capture rich context information. However, these methods use the same weights of feature detectors at all locations, and cannot robustly handle the variance of medical images.
\begin{figure*}[t]
\centering
\includegraphics[width=0.86\textwidth]{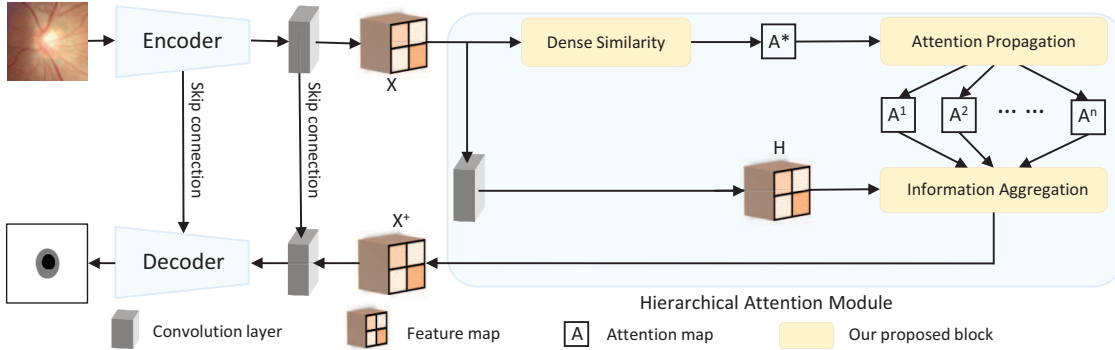}
\caption{The overall structure of our proposed Hierarchical Attention network. An input image is passed through an encoder and a bottleneck layer to produce a feature map $\mathbf{X}$. Then $\mathbf{X}$ is fed into the Hierarchical Attention module to reinforce the feature representation by our proposed Dense Similarity block, Attention Propagation block, and Information Aggregation block. Finally, the reinforced feature map $\mathbf{X}^+$ is transformed by a bottleneck layer and a decoder to generate the final segmentation results. The corresponding low-level and high-level features are connected by skip connections.}
\label{fig:model}
\end{figure*}

\textbf{Self-attention.} Self-attention has been widely used for natural language processing \cite{translate,transformer} and computer vision \cite{non-local}. Self-attention based methods aggregate global context information with dynamic weights, where they represent the context information with a weighted summation of information at all positions. DANet \cite{danet} models the attention in spatial and channel dimensions respectively. CCNet \cite{ccnet} obtains context information via an effective crisscross attention module. EMANet \cite{ema} reformulates the self-attention mechanism into an expectation-maximization iteration manner. Besides, Graph Convolutional Networks \cite{gcn} perform a message-passing similarly as self-attention. Curve-GCN \cite{curve-gcn} represents the object as a graph and conducts information fusion using a Graph Convolutional Network \cite{gcn}. Recently proposed high-order Graph Convolution methods \cite{high-order,mixhop,high-order-nn} also inspire us, which can learn a general class of neighborhood mixing relationships.

\textbf{Hierarchical attention.} Hierarchical attention has shown advantages in many tasks, such as document classification \cite{hie-doc}, response generation \cite{hie-respnse}, sentiment classification \cite{hie-senti}, action recognition \cite{hie-action}, and reading comprehension \cite{hie-read}. These methods are different from ours that obtains hierarchical attention via the high-order graph.
\section{Hierarchical Attention Network}
In the self-attention method, the updated features will contain much information about other categories, because of the effects of the inter-class indistinction, high variability, and noise in medical images. To address this issue, we explore a novel attention mechanism from the view of the high-order graph, which could reduce the noise in the attention map and adaptively aggregate global context information of multi-level. 

As illustrated in Figure \ref{fig:model}, we propose a Hierarchical Attention Network (HANet) for medical image segmentation. Constructed with an encoder-decoder architecture, HANet contains a hierarchical attention $\left(\mathbf{HA}\right)$ module to capture global context information over local features. First, a medical image is encoded by an encoder to form a feature map with the spatial size ($h\times w$). Then, after transformed by a channel reduction layer, the produced feature map $\mathbf{X} \in \mathbb{R} ^{c \times h \times w}$ is fed into the $\mathbf{HA}$ module to reinforce the feature representation by our hierarchical attention strategy. In the $\mathbf{HA}$ module, $\mathbf{X}$ is transformed by two branches. 1) the Dense Similarity block and the Attention Propagation block transform $\mathbf{X}$ orderly to generate initial attention map $\mathbf{A}^*$ and hierarchical attention maps $\left\{\mathbf{A}^h \mid h \in \left\{1, 2, ..., n\right\} \right\}$. 2) $\mathbf{X}$ is transformed by a bottleneck layer to produce feature map $\mathbf{H}$. Further, the feature map $\mathbf{H}$ and hierarchical attention maps $\left\{\mathbf{A}^h \mid h \in \left\{1, 2, ..., n\right\} \right\}$ are utilized to mix context information of multiple levels to produce new feature map $\mathbf{X}^+$ via the Information Aggregation block. Finally, the feature map $\mathbf{X^+}$ is transformed by a bottleneck layer and a decoder to generate the final segmentation results. Moreover, the low-level and high-level features are connected by skip connections~\cite{resnet}, which has been widely proved to recover the segmentation details successfully. Therefore, coupled with the $\mathbf{HA}$ module, HANet can capture richer context information and enhance the discriminative ability of feature representations. Subsequently, we introduce the hierarchical attention module in detail.
\subsection{Dense Similarity}
\begin{figure}[t]
\centering
\includegraphics[width=0.43\textwidth]{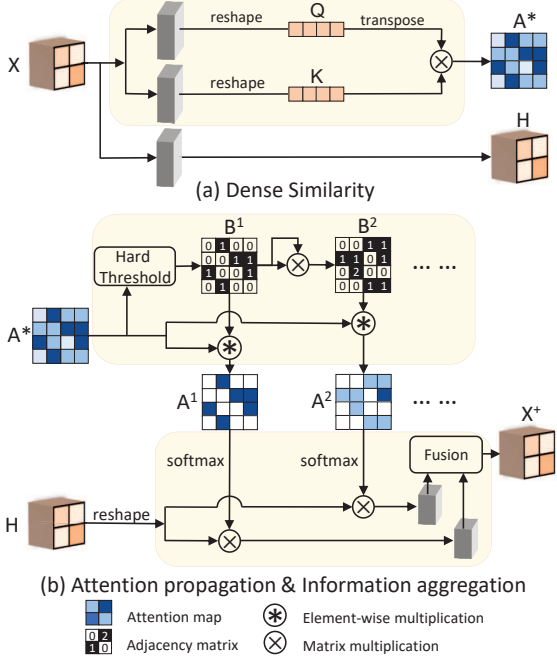}
\caption{The details of our Dense Similarity block, Attention Propagation block, and  Information Aggregation block.}
\label{fig:attention}
\end{figure}

The dense similarity block computes initial attention map $\mathbf{A}^*$, which represents the similarities between two corresponding features. As shown in Figure \ref{fig:attention} (a), we calculate $\mathbf{A}^*$ from $\mathbf{X}$ in a dot-product manner \cite{non-local, transformer}. Given the feature $\mathbf{X}$, we feed it to two parallel $1 \times 1$ convolutions to generate two new feature maps with shape $c \times h \times w$. Then both of them are reshaped to $\mathbb{R} ^{c \times (h \times w)}$ , namely $\mathbf{Q}$ and $\mathbf{K}$. The initial attention map $\mathbf{A}^* \in \mathbb{R} ^{(h \times w) \times (h \times w)}$ is calculated by matrix multiplication of $\mathbf{Q}^T$ and $\mathbf{K}$, as follows: 
\begin{equation}
\mathbf{A}^* = \alpha \mathbf{Q}^T \mathbf{K}
\label{eq:pcc}
\end{equation}
where $\alpha$ is a scaling factor to counteract the effect of the numerical explosion. Following previous work \cite{transformer}, we set $\alpha$ as $\frac{1}{\sqrt{c}}$ and $c$ is the channel number of $\mathbf{K}$. 

Computing the similarity between features in a dot-product manner is much faster and more space-efficient in practice. The initial attention map $\mathbf{A}^*$ is further utilized to form hierarchical attention maps in the next sections.
\subsection{Attention Propagation}
Our attention propagation block that produces the hierarchical attention maps $\left\{\mathbf{A}^h \mid h \in \left\{1, 2, ..., n\right\} \right\}$ from $\mathbf{A}^*$ is based on some basic theories of the graph. A graph is given in the form of adjacency matrix $\mathbf{B}^h$, where $ \mathbf{B}^h \left[i , j\right]$ is a positive integer if vertex $i$ can reach vertex  $j$  after $h$ hops, otherwise $ \mathbf{B}^h \left[i , j\right]$ is zero. And $\mathbf{B}^h$ can be computed by the adjacency matrix $\mathbf{B}^1$ multiplied by itself $h-1$ times:
\begin{equation}
\mathbf{B}^h =  \underbrace{\mathbf{B}^1 \mathbf{B}^1~...~...~\mathbf{B}^1}_h
\label{bh}
\end{equation}
If $\mathbf{B}^h$ is normalized into a bool accessibility matrix, where we set zero to false and others to true, then as $h$ increases, $\mathbf{B}^h$ tends to be equal to $\mathbf{B}^{h-1}$. At this point, $\mathbf{B}^h$ is the transitive closure\footnote{\url{https://en.wikipedia.org/wiki/Transitive_closure}} of the graph, where there is a direct edge between vertex $i$ and vertex $j$ if vertex $j$ is reachable from vertex $i$. Moreover, if we initialize an edge between two vertices only if they have the same label, vertices with the same label will form a complete graph\footnote{\url{https://en.wikipedia.org/wiki/Complete_graph}}. The transitive closure is the best case for attention mechanism, where the updated feature of node mixes the latent information of all features that have the same label.

Based on the above graph theories, we propose a novel way to realize the propagation of attention. As presented previously, the element of $\mathbf{A}^*$ represents the correlation between two corresponding features. We consider $\mathbf{A}^*$ as the adjacency matrix of a graph, where an edge means the degree that two nodes belong to the same category. As shown in the upper part of Figure~\ref{fig:attention} $\left(b\right)$, given $\mathbf{A}^*$, we erase those low confidence edges to produce the down-sampled graph $\mathbf{B}^1$ via applying a hard threshold. The operation is related to previous works \cite{region_mining, object_localization} which operate on the activation maps. We expect each node to be connected only to the nodes with the same label and to reduce the noise in the attention map. We obtain the down-sampled graph $\mathbf{B}^1$ as follows: 
\begin{equation}
\mathbf{B}^1 \left[i, j \right] =
\begin{cases}
1& \mathbf{A}^*\left[i, j \right] \geq \delta\\
0& \mathbf{A}^*\left[i, j \right] <  \delta 
\end{cases}
\label{eq:bl}
\end{equation}
where $\delta$ is a hard threshold. Then high-order graph $\mathbf{B}^h$ can be obtained with Eq.\ref{bh}, which indicates the $h$-degree neighbors of each node. Finally, the hierarchical attention maps are calculated as follows:
\begin{equation}
\mathbf{A}^h\left[i, j\right] = \mathbf{A}^*\left[i, j\right] \times  \mathbf{B}^h\left[i, j\right]
\label{eq:ah}
\end{equation}
where $h$ is an integer adjacency power indicating the steps of attention propagation. Thus the attention information of different levels is decoupled into different attention maps via $\mathbf{B}^h$. The produced hierarchical attention maps $\left\{\mathbf{A}^h \mid h \in \left\{1, 2, ..., n\right\} \right\}$ are used to aggregate hierarchical context information in section \ref{aggregate}. 

In our attention propagation method, the high-order graph $\mathbf{B}^h$ is important for building hierarchical attention maps because it can reduce noise in the initial attention map. Meanwhile, we can also realize our method without $\mathbf{B}^h$. Concretely, the hierarchical attention maps can also be obtained as follows:
\begin{equation}
\mathbf{A}^h =  \underbrace{\mathbf{A}^* \mathbf{A}^*~...~...~\mathbf{A}^*}_h
\end{equation}
where $\mathbf{A}^*$ is the initial attention map. To a certain extent, the above operation is related to the high-order Graph Convolution methods \cite{high-order,mixhop,high-order-nn}.
But in this way, the case of transitive closure mentioned above is difficult to be achieved, and the obtained hierarchical attention maps will result in unsuitable features that mix a large amount of context information from other categories. Thus it is inconsistent with our goal of reducing noise in the attention map. Furthermore, the meaning of $\mathbf{A}^h$ is very complex and hard to be explained in practice. Therefore, we propose the attention propagation method that produces hierarchical attention maps via the high-order graph $\mathbf{B}^h$. Our method reduces computational complexity and is clear to be interpreted.

\subsection{Information Aggregation}\label{aggregate}
As shown in the lower part of Figure \ref{fig:attention} (b), we aggregate information of multiple levels to generate $\mathbf{X}^+$ in a weighted sum manner, where we perform matrix multiplication between feature map $\mathbf{H}$ and the normalized hierarchical attention map $\tilde{\mathbf{A}^h}$, as follows:
\begin{equation}
\mathbf{X}^+ = \Gamma_\theta \left(\mathop{\begin{vmatrix} \\ \\ \end{vmatrix}}\limits_{h=1} \limits^{n}  W^h_\theta \left(\mathbf{H}\tilde{\mathbf{A}^h}\right) \right)
\label{eq:ia}
\end{equation}
where $
\tilde{\mathbf{A}^h}[i,j] = \frac{exp(\mathbf{A}^h[i,j])}{\sum _j exp(\mathbf{A}^h[i,j])}
$, and $\Vert$ denotes channel-wise concatenation, $n$ is the highest level of attention maps. $W^h_\theta$ and $\Gamma_\theta$ are $1 \times 1$ convolutions. Finally, $\mathbf{X}^+$ is transformed by a bottleneck layer and enriched by the decoder to generate accurate segmentation results.

Generally, our Hierarchical Attention module can be flexibly embedded in the existing fully convolutional networks, and it only mixes the context information of high correlation features thus enhances the discriminative ability of feature representations.

\section{Experiments}
To evaluate the proposed method, we carry out comprehensive experiments on three medical image segmentation tasks (i.e. optic disc/cup segmentation, retinal blood vessel segmentation, and lung segmentation). They correspond to three representative characteristics of the medical image (i.e. the inter-class indistinction, high variability, and noise). For optic disc/cup segmentation, we conduct experiments on the REFUGE dataset and Drishti-GS1 \cite{DGS} dataset. For blood vessel segmentation and lung segmentation, we conduct experiments on the DRIVE \cite{DRIVE} dataset and LUNA dataset respectively. 

\subsection{Datasets}
\textbf{REFUGE.} The dataset is arranged for the segmentation of optic disc and cup, which consists of 400 training images and 400 validation images. The testing set is not available. The training images are captured with the Zeiss Visucam 500 fundus camera at a resolution of 2124$\times$2056 pixels. The validation images are captured with the Canon CR-2 fundus camera at a resolution of 1634$\times$1634 pixels. The pixel-wise disc and cup gray-scale annotations are provided.

\textbf{Drishti-GS1.} It contains 50 training images and 51 testing images for optic disc/cup segmentation. All images are taken centered on optic disc with a field-of-view of 30 degrees and of dimensions 2896$\times$1944 pixels. The annotations are provided in the form of average boundaries.

\textbf{DRIVE.} The dataset is arranged for blood vessel segmentation. It includes 40 color retina images of dimensions 565$\times$584 pixels, from which 20 samples are used for training and the remaining 20 samples for testing. Manual annotations are provided by two experts, and the annotations of the first expert are used as the gold standard.

\textbf{LUNA.} It contains 2D CT images of dimensions 224$\times$224 pixels from the Lung Nodule Analysis (LUNA) competition which can be freely downloaded from the website\textsuperscript{\ref {LUNA}}. Following the previous work \cite{ce-net}, we use $80\%$ of the total 267 images for training and the rest for testing.
\subsection{Implementation Details}
We build our networks with PyTorch and train it on a single TITAN XP GPU. We choose the ImageNet \cite{imagenet} pre-trained ResNet-101 as our encoder designed in a fully convolutional fashion~\cite{fcn} that replaces the convolutions within the last blocks by dilated convolutions \cite{overfeat, local-global, dilated_conv}. Our encoder can adopt three different output strides (i.e. 16, 8, 4) for feature extraction. The setting of dilated convolutions is the same as ~\cite{deeplabv3} when output stride is 16 or 8, and we change the stride of the first convolution of ResNet-101 from 2 to 1 when output stride is 4. In the decoder, the input is first bilinearly upsampled and then fused with the corresponding low-level features. We adopt a simple yet effective decoder module as \cite{deeplabv3+} that only takes the output of the first block of ResNet-101 \cite{resnet} as low-level features. Finally, the output of the decoder is bilinearly upsampled to the same size as the input image, and we compute loss via cross-entropy loss function. After initializing two hyper-parameters of $\delta$ and $h$, our HANet is trained end-to-end. In addition, we implement DeepLabv3+ \cite{deeplabv3+} and DANet \cite{danet} for better comparison.

We use Stochastic Gradient Descent with mini-batch for training. The initial learning rate is 0.01. And we use momentum of 0.9 and weight decay of 5e-4. For optic disc/cup segmentation and lung segmentation, we set training time to 100 epochs and employ a learning rate policy of Reduce LR On Plateau where the learning rate is multiplied by 0.1 if the performance on validation set has no improvement within the previous five epochs. Moreover, the input spatial resolution is 513$\times$513 and the output stride is 16. For blood vessel segmentation, we set training time to 20 epochs and the learning rate is multiplied by 0.5 in the 10th epoch. The input spatial resolution is 224$\times$224 and the output stride is 8. We apply photometric distortion, rotation, random scale cropping, left-right flipping, and gaussian blurring during training for data augmentation.

\subsection{Results on Optic Disc/Cup Segmentation}
Optic Disc/Cup Segmentation is very useful in clinical practice and diagnosis of glaucoma, where glaucoma is usually characterized by the larger cup to disc ratio (CDR).  However, its segmentation is challenging because of the high similarity among the cup, disc, and background. We first conduct ablation experiments with different attention module settings on the REFUGE dataset. Then, the proposed method is compared with existing state-of-the-art segmentation methods. We also evaluate the robustness of HANet for domain adaptation, where the model trained only on the REFUGE training set is tested on the Drishti-GS1 dataset. For these experiments, we first localize the disc following the existing methods \cite{M-net, ET-Net, posal} and then transmit the cropped images into our network. Because the official testing set is not available, we use 50 images of the training set to select the best model and then test the best model with the official verification set. We do not perform any post-processing and the results are represented by the dice coefficient (Dice) and mean absolute error of the cup to disc ratio (E$_{CDR}$), where Dice$_d$ and Dice$_c$ denote dice coefficients of optic disc and cup respectively.
\begin{figure}[t]
\centering
\includegraphics[width=0.43\textwidth]{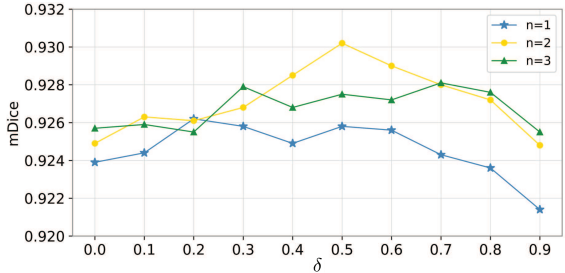}
\caption{The influence of the threshold $\delta$ and adjacency power $h$, and the results are represented by mDice on the REFUGE validation set. $n$ is the highest level of attention, where $h \in \left\{1, 2, ..., n\right\}$.}
\label{fig:hop} 
\end{figure}
\subsubsection{Ablation Study for Attention Modules}
We study the influence of the threshold $\delta$ and adjacency power $h$ on our hierarchical attention module. After normalizing the initial attention map between 0 and 1, we use threshold $\delta$ to get the down-sampled graph as Eq.\ref{eq:bl}. And $h$ is used in Eq.\ref{bh}. As shown in Figure \ref{fig:hop}, HANet is not highly sensitive for different parameter configurations. Specifically, the line of $n=1$ denotes there is only one level attention, which shows a gradual increase when $\delta <= 0.3$ and a downward trend when $\delta > 0.3$. It means that the part of the noise in the attention map is erased by a small threshold and the important attention information is discarded by a big threshold. HANet performs better when $n>1$, where there are more than one level of attention map. For the line of $n=2$, HANet achieves the best performance when setting $\delta = 0.5$. When $\delta > 0.7$, the results of $n=3$ outperforms the results of $n<3$. It reveals the attention propagation module with a larger $n$ can infer from existing attention to more positions, so HANet still performs well even if most of the attention information is discarded by a large threshold. These two parameters enable our model to learn a wider range of feature representations and to robustly adapt to many tasks.
\begin{table}[t]
\centering
\begin{tabular}{l|c|c|c|c} 
\hline
Method & mDice & E$_{CDR}$ & Dice$_{d}$ & Dice$_{c}$ \\
\hline
AIML\textsuperscript{\ref {REFUGE_val}} & 0.9250 & 0.0376 & 0.9583 & 0.8916\\
BUCT \textsuperscript{\ref {REFUGE_val}} &0.9188 &0.0395 & 0.9518 & 0.8857\\
CUMED\textsuperscript{\ref {REFUGE_val}} &0.9185 & 0.0425 & 0.9522 & 0.8848\\
VRT\textsuperscript{\ref {REFUGE_val}} &0.9161 & 0.0455 & 0.9472 & 0.8849\\
CUHKMED\textsuperscript{\ref {REFUGE_val}} & 0.9116 & 0.0440 & 0.9487 & 0.8745 \\
\hline
U-Net\cite{U-net}& 0.8926& - &0.9308 & 0.8544 \\
POSAL\cite{posal} &0.9105 & 0.0510 & 0.9460 &0.8750 \\
M-Net\cite{M-net}& 0.9120 & 0.0480 &0.9540 &0.8700 \\
Ellipse\cite{ellipse} &0.9125 & 0.0470 & 0.9530 & 0.8720 \\
Task-DS\cite{task-ds}& 0.9129& - & - & - \\
ET-Net\cite{ET-Net}& 0.9221 & - & 0.9529 & 0.8912\\
DeepLabv3+\cite{deeplabv3+}& 0.9215 & 0.0403 & 0.9575 & 0.8854\\
DANet\cite{danet}&0.9192 & 0.0423 & 0.9572 & 0.8813 \\
\hline
HANet$_{h1}$ & 0.9239 & 0.0355 & 0.9544 & 0.8934\\
HANet$_{h2}$ & \textbf{0.9302} & \textbf{0.0347} & \textbf{0.9599} & \textbf{0.9005}\\
\hline
\end{tabular}
\caption{Optic disc/cup segmentation results on REFUGE validation set. HANet$_{h1}$ indicates that HANet is set to $\delta = 0$ and $h \in \left\{1\right\}$, which recovers the one-level attention. For HANet$_{h2}$, we set $\delta = 0.5$ and $h \in \left\{1, 2\right\}$.}
\label{table:refuge}
\end{table}

\begin{figure*}[ht]
\centering
\includegraphics[width=0.86\textwidth]{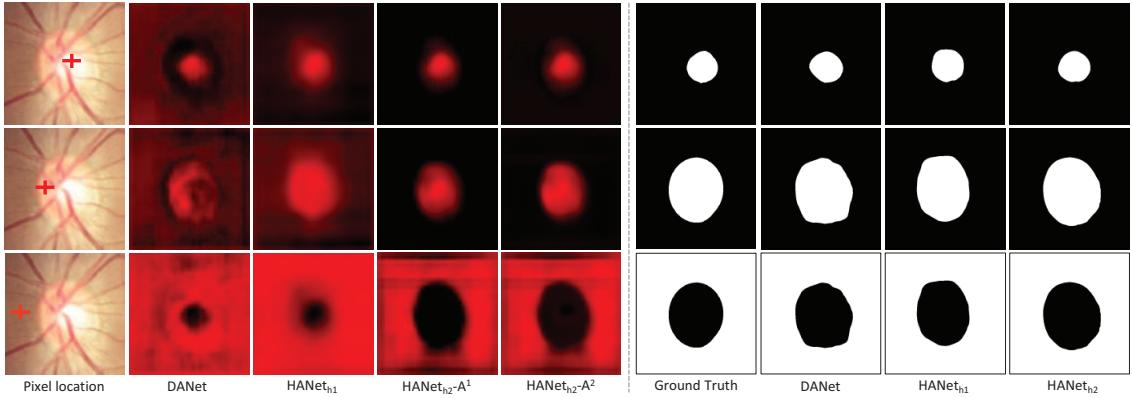}
\caption{Visualization of the attention maps of pixel marked by ``+'' and segmentation results on REFUGE validation set. HANet$_{h1}$ indicates setting our HANet to $\delta = 0$ and $h \in \left\{1\right\}$, which recovers the original self-attention method. For our HANet$_{h2}$, we set $h \in \left\{1, 2\right\}$, which generates two levels of attention map (i.e., $A^1$ and $A^2$). The second to fifth columns show the attention maps generated by different models and the sixth to ninth columns show the segmentation results. Obviously, there is much noise in the attention maps of the second column to the third column, which is consistent with our assumption that the attention maps produced by the original self-attention based method contain noise. Best viewed in color.}
\label{fig:map}
\end{figure*}

\subsubsection{Comparing with State-of-the-art }
We compare our HANet with existing methods on the REFUGE\textsuperscript{\ref {REFUGE}} set. Our HANet$_{h2}$ ($\delta = 0.5$ and $h \in \left\{1, 2\right\}$) is compared with two baselines namely DeeplabV3+ \cite{deeplabv3+} and DANet \cite{danet}, where Deeplabv3+ \cite{deeplabv3+} has a similar encoder-decoder structure to our HANet, and DANet \cite{danet} models the attention in spatial (one-level self-attention) and channel dimensions respectively. To compare the multi-level and one-level attention fairly, we also compare HANet$_{h2}$ with HANet$_{h1}$ ($\delta = 0$ and $h \in \left\{1\right\}$), where HANet$_{h1}$ recovers the original self-attention method. HANet$_{h2}$ is also compared with methods leading the REFUGE challenge\footnote{\url{https://refuge.grand-challenge.org/Results-ValidationSet_Online/}\label{REFUGE_val}} in conjunction with MICCAI 2018 (e.g. AIMI, BUCT). 

As shown in Table \ref{table:refuge}, our HANet$_{h2}$ achieves the best performance among the competitive published benchmarks. In particular, HANet$_{h2}$ outperforms Deeplabv3+ \cite{deeplabv3+} and DANet \cite{danet} by $0.87\%$ and $1.1\%$ on mDice respectively, and it also outperforms the previous state-of-the-art optic disc/cup segmentation method ET-Net \cite{ET-Net}. Moreover, our HANet$_{h2}$ outperforms the AIML\textsuperscript{\ref {REFUGE_val}} which achieved the first place for the optic disc and cup segmentation tasks in the REFUGE challenge. Notably, our model achieves impressive results for optic cup segmentation, which is an especially difficult task because of the high similarity between optic cup and disc. 

Figure \ref{fig:map} shows the attention maps of ``+'' marked pixels. In the self-attention based methods, the feature of ``+'' marked pixel will be updated via the weighted summation of features in other locations, where weights are represented by the red color shown in the attention maps. Obviously, the attention maps from the second column to the third column contain noise, which may result in the weighted summed features contain much information about other categories. As shown in the fourth and fifth columns of Figure \ref{fig:map}, our HANet$_{h2}$ that aggregates context information with hierarchical attention only focuses on the positive context information, which means our approach naturally reduces the noise in the attention map. 
\begin{table}[t]
\centering
\begin{tabular}{l|c|c|c|c} 
\hline
Method & mDice & E$_{CDR}$ & Dice$_d$ & Dice$_c$ \\
\hline
M-Net\cite{M-net}& 0.8515 & 0.1660 &0.9370 &0.7660 \\
Ellipse\cite{ellipse} &0.8520 & 0.1590 & 0.9270 & 0.7770 \\
DeepLabv3+\cite{deeplabv3+}&0.8643 & 0.1781 & 0.9663 & 0.7623 \\
DANet\cite{danet} &0.8924 & 0.1131 & 0.9660 & 0.8187 \\ 
\hline
HANet$_{h1}$ & 0.8930 & 0.1346& 0.9629 & 0.8232\\
HANet$_{h2}$ & \textbf{0.9117} &\textbf{0.1091} & \textbf{0.9721} & \textbf{0.8513}\\
\hline
\end{tabular}
\caption{The comparison of the robustness of different models on Drishti-GS1 dataset. We use the training set of REFUGE dataset to train the model and test the trained model with the whole Drishti-GS1 dataset.}
\label{table:dgs}
\end{table}
\subsubsection{Robustness} 
We report the results of HANet on the Drishti-GS1 dataset \cite{DGS} to evaluate its robustness for domain adaptation, where the model is only trained on the REFUGE training set. Domain adaptation is a great challenge for the medical image segmentation because the diversity of shooting equipment dramatically affects the quality of images. As shown in Table \ref{table:dgs}, the proposed HANet$_{h2}$ achieves the best robust performance than other state-of-the-art segmentation methods. In particular, our HANet$_{h2}$ that aggregates context information with dynamic weights outperform DeepLabv3+ \cite{deeplabv3+} which aggregates context information with fixed weights by $4.74\%$ on mDice. Moreover, Our HANet$_{h2}$ that aggregates context information with hierarchical attention outperforms the HANet$_{h1}$ and DANet \cite{danet} which aggregates context information with one-level attention by $1.87\%$ and $ 1.92\%$ on mDice, respectively. Especially, our HANet$_{h2}$ outperforms state-of-the-art Ellipse \cite{ellipse} by $7.43 \%$ on Dice$_c$, which illustrates that our hierarchical attention has great advantages in the recognition of confusing categories. The above results demonstrate our method is more robust to the variance of input than existing methods.

\begin{table}[t]
\centering
\begin{tabular}{l|c|c|c|c} 
\hline
Method & ACC & F1 & Se  & Sp \\
\hline
DeepVessel\cite{deepvessel}&0.9523 &0.7900 &0.7603  & - \\
U-net\cite{U-net}          &0.9554 & 0.8175& 0.7849 & 0.9802\\
R2U-net\cite{r2u-net}      &0.9556 &0.8171 & 0.7792 &0.9813\\
LadderNet\cite{laddernet}  &0.9561 &0.8202 & 0.7856 &0.9810\\
Multi-scale\cite{ms-nfn}        &0.9567 & -     &0.7844  & 0.9819\\
DRIU\cite{driu}            & -&  0.8221 & 0.8264 & - \\
Vessel-Net\cite{vessel-net}& 0.9578 &- &0.8038 & 0.9802\\
DUNet\cite{dunet}          & 0.9566 & 0.8237 & 0.7963 &  0.9800\\
DEU-Net\cite{deu-net}      & 0.9567& 0.8270 & 0.7940 &0.9816\\
CASR\cite{casr} &-      &\textbf{0.8353} & \textbf{0.8419}  & -\\
DeepLabv3+\cite{deeplabv3+}&0.9679  &0.8222  &0.7881  & 0.9814\\
DANet\cite{danet}& 0.9693 & 0.8205 & 0.7827 & 0.9804\\
\hline
HANet$_{h1}$ &\textit{0.9706} &0.8251 & 0.8158& \textit{0.9832} \\
HANet$_{h2}$ & \textbf{0.9712} & \textit{0.8300}&  \textit{0.8297}& \textbf{0.9843} \\
\hline
\end{tabular}
\caption{Blood vessel segmentation results on DRIVE testing set.}
\label{table:DRIVE}
\end{table}

\begin{figure*}[ht]
\centering
\includegraphics[width=0.86\textwidth]{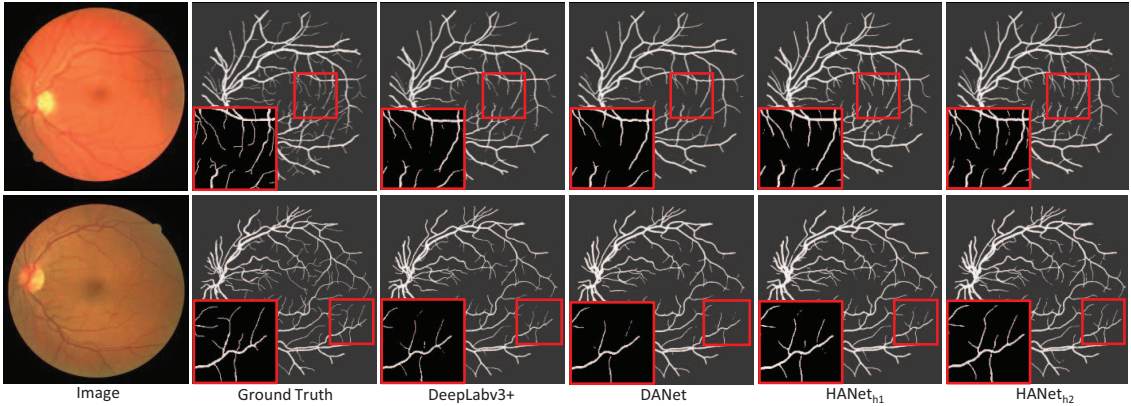}
\caption{Visualization results on DRIVE testing set. The local patches is highlighted for a more detailed comparison.}
\label{fig:map3}
\end{figure*}

\subsection{Results on Retinal Blood Vessel Segmentation}
Retina blood vessel is a typical object with curvilinear structure, and its segmentation is challenging because the blood vessel is too small and diverse in shape. Our method is compared with the DeepLabv3+ \cite{deeplabv3+}, DANet \cite{danet}, and other state-of-the-art vessel segmentation methods. Following the existing methods \cite{vessel-net, laddernet}, we orderly extract $48 \times 48$ patches with a stride of 8 along with both horizontal and vertical directions for training, and we recompose the entire image with probability maps of partly overlapped patches in the testing phase. We do not perform any post-processing and the results are represented by the accuracy (\textbf{ACC}), F1-score (\textbf{F1}), sensitivity (\textbf{Se}), and specificity (\textbf{Sp}).

Extensive results on the DRIVE testing set are shown in Table \ref{table:DRIVE}. HANet$_{h2}$ attains the highest values on \textbf{ACC} and \textbf{Sp} while the values of other two metrics are still competitive to the Context-aware Spatio-recurrent (CSAR) method \cite{casr}. CSAR \cite{casr} only handles the curvilinear structure segmentation, but our method could segment many kinds of medical images. Further, results also show that HANet$_{h2}$ outperforms Deeplabv3+ \cite{deeplabv3+} and one-level attention based methods (i.e. HANet$_{h1}$ and DANet \cite{danet}). Qualitative results are shown in Figure \ref{fig:map3}, which demonstrates the better effectiveness of our HANet to detect the small and high variable object.
\begin{table}[t]
\centering
\begin{tabular}{l|c|c|c|c} 
\hline
Method & IoU & ACC & F1  & Se \\
\hline 
U-Net \cite{U-net} & 0.9130 & 0.9750 & - & 0.9380 \\
ResU-Net\cite{r2u-net} & - & 0.9849 & 0.9690 & 0.9555 \\
RU-Net\cite{r2u-net} & - &0.9836 & 0.9638 & 0.9734 \\
R2U-Net\cite{r2u-net}& - &0.9918 & 0.9823 &0.9832 \\
CE-Net\cite{ce-net}&0.9620 & 0.9900 & - & 0.9800\\
DeepLabv3+\cite{deeplabv3+}&0.9674 &0.9923  &0.9834  &0.9851 \\
DANet\cite{danet}&0.9593 & 0.9903 & 0.9792 & 0.9832 \\
\hline
HANet$_{h1}$&0.9662  &0.9920 & 0.9828&0.9842 \\
HANet$_{h2}$& \textbf{0.9768}  & \textbf{0.9945} &  \textbf{0.9883}& \textbf{0.9879} \\
\hline
\end{tabular}
\caption{Experimental results on LUNA testing set.}
\label{table:LUNA}
\end{table}

\begin{figure}[t]
\centering
\includegraphics[width=0.43\textwidth]{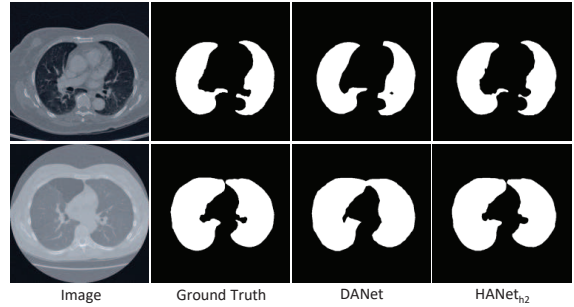}
\caption{Visualization results on LUNA testing set.}
\label{fig:lung}
\end{figure}
\subsection{Results on Lung Segmentation}
We apply HANet to segment lung structure in 2D CT images, where noise is closely related to image quality. The segmentation results are represented by the accuracy (\textbf{ACC}), Intersection over Union (\textbf{IoU}), F1-score (\textbf{F1}), and sensitivity (\textbf{Se}). As shown in Table \ref{table:LUNA}, the HANet$_{h2}$ achieves state-of-the-art performance in all metrics. Further, two example segmentation results are shown in Figure \ref{fig:lung}. The above results show our hierarchical attention module substantially boosts the segmentation performance in medical images which are easily infected by noise.
\section{Conclusion}
In this paper, we experimentally find that the attention map contains a lot of noise in the original self-attention method, which leads to that the updated feature mixes much context information about other categories. The noise in the attention map has a serious impact on the feature representation of the medical images with inter-class indistinction. And we propose a novel Hierarchical Attention Network (HANet) for medical image segmentation, which adaptively captures multi-level global context information in a high-order graph manner. Especially, the hierarchical attention module embedded in the HANet can be flexibly inserted into existing CNNs.  
Extensive experiments demonstrate that our hierarchical attention module naturally reduces noisy attention information caused by inter-class indistinction, and it is more robust to the variance of input than the original self-attention based methods. Our HANet achieves outstanding performance consistently on four benchmark datasets. 
{\small
\bibliographystyle{ieee_fullname}
\bibliography{egbib}
}

\end{document}